\documentclass[10pt,twocolumn,letterpaper]{article}

\usepackage[pagenumbers]{cvpr} % To force page numbers, e.g. for an arXiv version

\usepackage{booktabs}
\usepackage{multirow}
\usepackage{graphicx}
\usepackage{colortbl}
\usepackage{pifont}

\definecolor{tabfirst}{rgb}{1, 0.7, 0.7}
\definecolor{tabsecond}{rgb}{1, 0.85, 0.7}
\definecolor{tabthird}{rgb}{1, 1, 0.7}

\newcommand{\cmark}{\ding{51}}
\newcommand{\xmark}{\ding{55}}

\definecolor{cvprblue}{rgb}{0.21,0.49,0.74}
\usepackage[pagebackref,breaklinks,colorlinks,allcolors=cvprblue]{hyperref}

\def\confName{CVPR}
\def\confYear{2027}

\title{TurboWAM: Learning What Matters for Fast World Action Models}

\author{Guile Wu$^1$, Shuangzhi Li$^{1,3,}${\footnotemark[1]}~, Xingxin Chen$^1$, Chengjie Huang$^1$, Sam Hosseini$^1$,\\
Tongtong Cao$^2$, Dongfeng Bai$^1$, and Yingxue Zhang$^1$ \\
$^1$Huawei Noah's Ark Lab \quad $^2$Department of Foundation Model, Huawei \quad $^3$University of Alberta \\
\texttt{\url{https://turbowam.github.io/}}
}

\begin{document}
% \maketitle

\twocolumn[{%
\renewcommand\twocolumn[1][]{#1}%
\maketitle
% \vspace{-1cm}
\begin{center}
    \centering
    \captionsetup{type=figure}
    \includegraphics[height=7.cm]{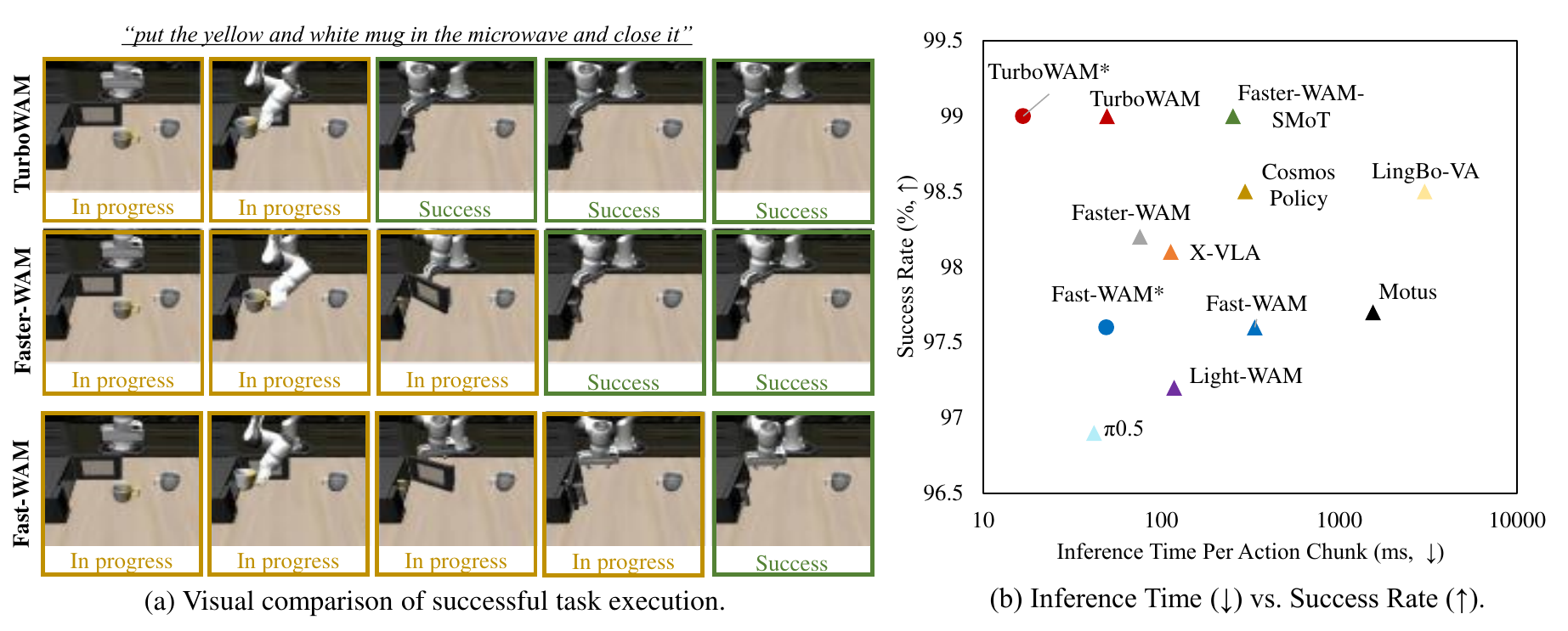}
    % \vspace{-0.3cm}
    \captionof{figure}{
        Overview of TurboWAM performance.
        (a) Visual comparison of successful task execution, highlighting the fast task completion of TurboWAM.
        (b) Comparison of inference time vs. success rate on LIBERO.
        TurboWAM achieves a better trade-off between inference time and success rate compared to state-of-the-art methods.
        TurboWAM* and Fast-WAM* denote the compiled versions.
        TurboWAM/TurboWAM* achieves 6.8$\times$/2.9$\times$ speedup compared to Fast-WAM/Fast-WAM*, respectively.
        }
    \label{fig:illustration}
\end{center}%
}]

\newcommand\blfootnote[1]{%
  \begingroup
  \renewcommand\thefootnote{}\footnote{#1}%
  \addtocounter{footnote}{-1}%
  \endgroup
}
\blfootnote{\textsuperscript{*}S. Li contributed to this work during an internship at Huawei Canada.}

\begin{abstract}
World action models have demonstrated promising performance in robot policy learning, yet their high inference latency remains a critical bottleneck for real-world deployment.
Recent approaches have sought to improve inference efficiency by eliminating test-time future-frame imagination or accelerating action prediction.
However, aggregating representations from all video world model layers can introduce redundant computation, while standard backpropagation may distribute action-supervision gradients unevenly across layers, limiting the effectiveness of world representation learning.
In this work, we propose TurboWAM, a fast world action model that learns to efficiently aggregate and optimize world representations for action prediction.
Specifically, TurboWAM introduces an efficient world representation aggregation mechanism that restricts aggregation to a bounded range of relevant world model layers identified through layer relevance analysis.
Within this bounded range, TurboWAM estimates aggregation weights for different world model layers based on the current action denoising state and combines them with learnable layer-wise weights to fuse world representations.
Moreover, TurboWAM employs a world model gradient modulation mechanism that decomposes action-supervision gradients into a layer-shared component and layer-specific residual components, enabling more balanced gradient propagation across world model layers.
This mechanism operates exclusively during training and introduces no additional inference computation.
Together, these designs enable more efficient world representation learning and low-latency action prediction.
Extensive experiments on multiple robot manipulation benchmarks and real-world manipulation tasks show that TurboWAM achieves low inference latency while maintaining competitive task success rates compared with state-of-the-art methods.
\end{abstract}   
\section{Introduction}
\label{sec:introduction}

World Action Models (WAMs)~\cite{ye2026world,kim2026cosmos,li2026causal,bi2026motus,yuan2026fast,zhang2026world} have recently emerged as a promising paradigm for robot policy learning.
Typically, WAMs integrate future visual generation with action prediction into a unified framework, allowing robots to anticipate how the physical world evolves during interaction.
Despite their promising performance, the high computational cost of WAMs remains a critical bottleneck for real-world deployment.
In particular, WAMs often rely on large-scale video generative models that require iterative denoising to predict future observations, resulting in substantial computational overhead and inference latency.
This limits their applicability to real-time robotic control, where efficient and responsive action prediction is essential.

Recently, several methods~\cite{yuan2026fast,ma2026faster,zhao2026faster,li2026efficient,li2026light} have been proposed to improve the inference efficiency of WAMs.
These methods primarily address the computational bottleneck by reducing test-time future visual imagination, accelerating action prediction, or developing more efficient model architectures.
Despite their encouraging results, several challenges remain.
First, aggregating representations from all video world model layers can incur redundant computation, as different layers contribute unequally to action prediction.
Moreover, the relevance of individual layers to action prediction may vary across denoising steps.
Second, when action supervision is backpropagated through the weighted aggregation, different world model layers receive imbalanced gradients according to their aggregation weights.
In particular, layers with smaller aggregation weights receive weaker action supervision, potentially limiting their optimization for world representation learning.

To address these challenges, in this work, we propose TurboWAM, a novel fast world action model that learns to efficiently aggregate and optimize world representations for action prediction.
Specifically, TurboWAM introduces an efficient world representation aggregation mechanism that restricts aggregation to a bounded range of relevant world model layers guided by layer analysis.
Within this bounded range, TurboWAM estimates aggregation weights for different world model layers based on the current action denoising state and further combines them with learnable layer-wise weights to fuse world representations.
Furthermore, TurboWAM develops a world model gradient modulation mechanism that decomposes action-supervision gradients into a layer-shared component and layer-specific residual components.
By combining these two components, the proposed mechanism promotes more balanced gradient propagation across the participating world model layers, facilitating their optimization for world representation learning.
Importantly, gradient modulation operates exclusively during training and introduces no additional inference computation.
Following~\cite{yuan2026fast,ma2026faster}, TurboWAM preserves joint video and action training while eliminating future visual prediction during inference.
During inference, only the bounded range of world model layers is executed once, and their key and value representations are cached and reused across action denoising steps, avoiding redundant backbone computation.
Together, these designs enable more efficient world representation utilization and low-latency action prediction.
We conduct extensive experiments on multiple robot manipulation benchmarks and on real-world robot manipulation tasks.
Experimental results demonstrate that TurboWAM achieves low inference latency while maintaining competitive task success rates compared with state-of-the-art methods (as illustrated in Figure~\ref{fig:illustration}).

In summary, our \textbf{contributions} are:
\begin{itemize}
    \item We propose TurboWAM, a fast world action model that improves inference efficiency through effective world representation aggregation and optimization while preserving the benefits of joint video and action training.
    \item We introduce an efficient world representation aggregation mechanism that reduces redundant computation and improves the utilization of world representations for action prediction.
    \item We develop a world model gradient modulation mechanism that promotes more balanced action supervision across world model layers, facilitating effective world representation learning.
    \item Extensive experiments on multiple robot manipulation benchmarks and real-world robot manipulation tasks demonstrate the effectiveness and efficiency of TurboWAM compared with state-of-the-art methods.
\end{itemize}

\begin{figure*}[t]
\centering
\includegraphics[width=1.0\linewidth]{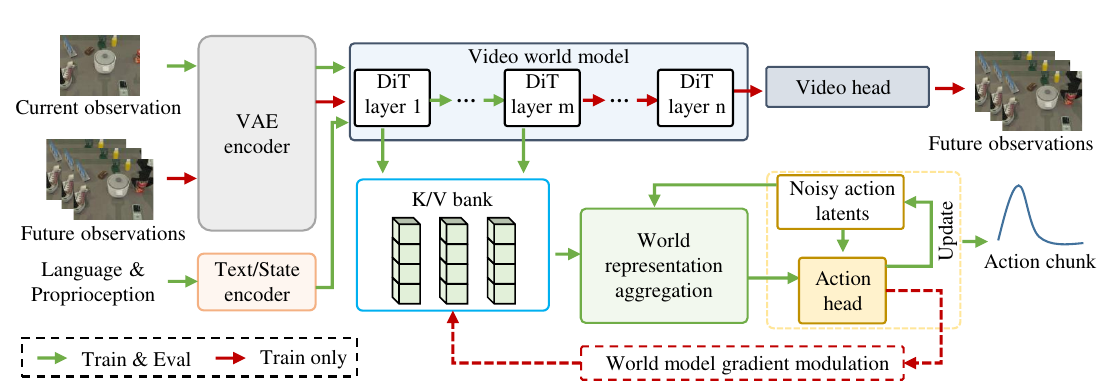}
\caption{Overview of the TurboWAM framework.
The world representation aggregation mechanism operates within a bounded range of relevant world model layers, estimating aggregation weights based on the current action denoising state and combining them with learnable layer-wise weights.
The world model gradient modulation mechanism decomposes and modulates action-supervision gradients across world model layers during the backward pass.
Together, these complementary designs improve world representation learning and reduce inference latency while maintaining competitive task success rates.
}
\label{fig:framework}
\end{figure*}
\section{Related Work}
\label{sec:related_work}

\paragraph{World Action Models.}
World Action Models (WAMs)~\cite{ye2026world,kim2026cosmos,zhang2026learning,zhu2025unified,zhang2026imagewam} and Vision-Language-Action (VLA) Models~\cite{black2024pi_0,black2025pi_05,kim2024openvla,kim2025fine} have emerged as two prominent paradigms for robot policy learning.
While VLA models typically predict actions directly from visual observations and language instructions, WAMs leverage visual generative models, such as video generation models~\cite{wan2025wan}, to incorporate knowledge of physical dynamics into action prediction.
By jointly modeling future visual observations and robot actions, WAMs learn world representations that capture how the environment evolves during interaction~\cite{ye2026world,kim2026cosmos,li2026causal,bi2026motus,yuan2026fast}.
Representative approaches, such as DreamZero~\cite{ye2026world}, Motus~\cite{bi2026motus}, Cosmos Policy~\cite{kim2026cosmos}, and LingBot-VA~\cite{li2026causal}, have demonstrated the potential of leveraging predictive world modeling for robot control.
Despite their promising performance, these methods often incur substantial computational overhead due to the expensive process of future visual generation, limiting their applicability to real-time robotic control.
Our work focuses on improving the inference efficiency of WAMs while maintaining competitive task success rates.

\vspace{-0.3cm}
\paragraph{Low-Latency World Action Models.}
Recent efforts have explored various strategies to reduce the inference latency of WAMs.
Fast-WAM~\cite{yuan2026fast} eliminates explicit future visual generation during inference while retaining video co-training during training, enabling efficient action prediction without test-time future visual imagination.
Faster-WAM~\cite{ma2026faster} aggregates key and value representations from all video world model layers using learned fusion weights and employs a single-DiT-block action head to reduce action prediction latency.
ARPS~\cite{gu2026right} aggregates intermediate world model features into a compact state for the action expert to reduce inference overhead and employs future-representation supervision to enhance world representation learning.
Other methods explore model compression and lightweight architectures~\cite{li2026efficient,li2026light}, as well as sparse video-action interactions~\cite{zhao2026faster}, to improve inference efficiency.
Despite these advances, aggregating representations from all world model layers can introduce redundant computation, as their contributions to action prediction may differ across denoising steps.
Furthermore, backpropagating action supervision through weighted aggregation may produce imbalanced gradients across world model layers.
To address these challenges, TurboWAM introduces a world representation aggregation mechanism to reduce redundant computation and improve representation utilization, and a world model gradient modulation mechanism to balance action-supervision gradients across layers and facilitate world representation learning.

\begin{figure*}[t!]
\centering
\includegraphics[width=1.0\linewidth]{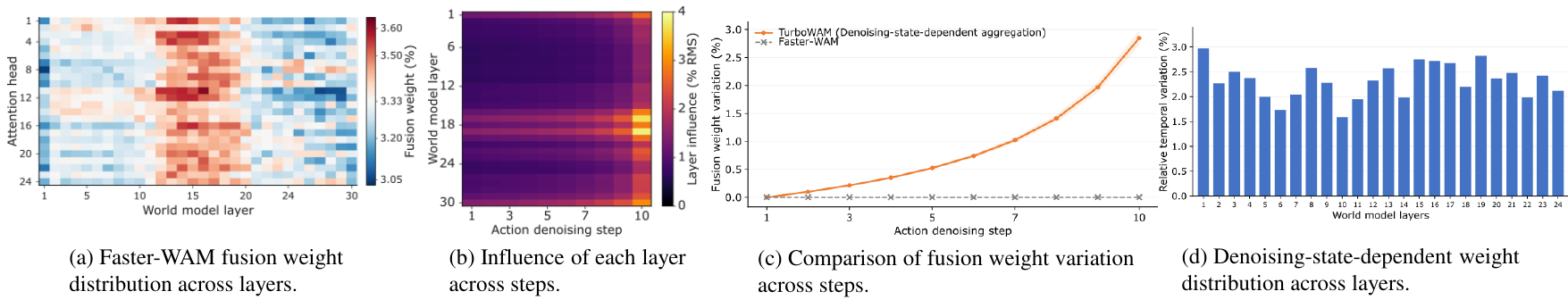}
\caption{Motivation for world representation aggregation in TurboWAM.
(a) Distribution of fusion weights across world model layers in Faster-WAM.
(b) Influence of individual world model layers at different denoising steps in Faster-WAM, measured by the relative change in predicted action velocity when the key/value pairs from a given layer are excluded from fusion.
(c) Comparison of fusion weights variation across action denoising steps between Faster-WAM and TurboWAM (denoising-state-dependent aggregation), measured by variation distance between layer fusion weight at each denoising step and its fusion weight at the first denoising step.
(d) Denoising-state-dependent fusion weight distribution across world model layers in TurboWAM, measured by fusion weight fluctuation relative to its mean across action denoising steps.
}
\label{fig:motivation}
\end{figure*}

\section{Methodology}
\label{sec:methodology}

\subsection{Approach Overview}
\label{sec:approach_overview}

\paragraph{Problem Formulation.}
This work focuses on world action models for robot control.
Given a current observation $o$, a language instruction $l$, and a proprioceptive state $p$, the goal is to predict an action chunk $a_{1:H}$ with horizon $H$.
In conventional WAMs, future visual imagination and action prediction are jointly modeled during inference, where the world model extracts representations of the current observation and predicts representations of future states:
\begin{equation}
\begin{aligned}
    z_f, z_o &= F_w(o,l,p),\\
    a_{1:H} &= F_a(z_f,z_o,l,p),
\end{aligned}
\end{equation}
where $z_f$ and $z_o$ denote the future and current observation representations, respectively, $F_w$ is the video world model for representation extraction (the variational autoencoder (VAE) and language and proprioception encoders are omitted for brevity), and $F_a$ is the action expert for action prediction.
Here, $a_{1:H}=(a_1,\ldots,a_H)$ denotes the predicted action chunk.
However, generating future observation representations incurs substantial computational overhead during inference.
Following~\cite{yuan2026fast,ma2026faster}, we decouple future visualimagination from action prediction, jointly optimizing video generation and action prediction during training while eliminating future visual prediction during inference.
The language instruction and proprioceptive state are encoded into world representations.

\vspace{-0.3cm}
\paragraph{Framework Overview.}
An overview of TurboWAM is shown in Figure~\ref{fig:framework}.
\emph{During training}, image observations are encoded using a VAE encoder, while language instructions and proprioceptive states are processed by their respective encoders to obtain latent representations.
Noise is added to the future-observation latents, which are fed into the video world model $F_w$ for denoising, conditioned on the current-observation latents, language instructions, and proprioceptive states.
All $n$ world model layers participate in video co-training, while the key and value representations from the first $m$ layers are used for action prediction.
Following~\cite{ma2026faster}, we employ a single DiT block as the action head, where noisy action latents provide queries that attend to the aggregated world representations together with the action-head representations.
The aggregation weights are estimated based on the current noisy action latents and further combined with learnable layer-wise weights.
During backward propagation, the proposed world model gradient modulation mechanism modifies action-supervision gradients propagated to the participating world model representations, while the video loss follows ordinary backpropagation.
\emph{During inference}, TurboWAM predicts actions from the current observation, language instruction, and proprioceptive state without performing future visual prediction.
Only the first $m$ world model layers are executed once, and their key and value representations are cached for reuse throughout action denoising.
The inference process is formulated as:
\begin{equation}
\begin{aligned}
    \mathcal{C}_m
    &= \{(K_i,V_i)\}_{i=1}^{m}
    = F_w^{(1:m)}(o,l,p),\\
    z_o^t
    &= \mathcal{A}(\mathcal{C}_m,a_{1:H}^{t}),\\
    a_{1:H}^{t-1}
    &= F_a(a_{1:H}^{t},z_o^t,t),
    \quad t=T,\ldots,1,
\end{aligned}
\label{eq:inference}
\end{equation}
where $F_w^{(1:m)}$ denotes the first $m$ layers of the video world model, $\mathcal{C}_m$ denotes their cached key and value representations, and $\mathcal{A}$ denotes the proposed world representation aggregation mechanism.
Here, $z_o^t$ represents the aggregated world representations at denoising step $t$, $a_{1:H}^{t}$ denotes the current noisy action latents, $a_{1:H}^{t-1}$ denotes the updated action latents after one denoising step, and $T$ is the total number of action denoising steps.
The action expert $F_a$ includes the action head and the sampler update, and the final predicted action chunk is $a_{1:H}=a_{1:H}^{0}$.
The cached representations remain unchanged throughout action denoising, while the aggregation weights and fused representations are updated at each denoising step.
World model gradient modulation is not applied during inference.

\subsection{World Representation Aggregation}
\label{sec:world_representation_aggregation}

\paragraph{Motivation.}
In WAMs, the video world model produces hierarchical representations across multiple layers, capturing different levels of abstraction and spatial information.
Fast-WAM~\cite{yuan2026fast} employs a fixed one-to-one correspondence between world model layers and action head layers without explicitly aggregating world representations across layers.
Faster-WAM~\cite{ma2026faster} instead employs a docking interface to aggregate key and value representations from all world model layers using learned fusion weights.
Although this provides a flexible way to utilize hierarchical world representations, it still requires executing the entire world model backbone during inference, introducing unnecessary computational overhead.
As shown in Figure~\ref{fig:motivation}(a), the docking interface of Faster-WAM assigns larger fusion weights to intermediate world model layers than to early and late layers, indicating that different layers contribute unequally to action prediction.
Figure~\ref{fig:motivation}(b) illustrates that the influence of individual world model layers on action prediction varies across denoising steps.
These observations motivate effective world representation aggregation that considers the contributions of relevant world model layers and their varying relationships with the current action denoising state.

\vspace{-0.3cm}
\paragraph{Aggregation Mechanism.}
Let $n$ denote the total number of layers\footnote{The terms transformer ``layer'' and ``block'' are used interchangeably.} in the world model, and let $K_i$ and $V_i$ denote the key and value representations of the current observation extracted from the $i$th layer, respectively.
Based on the layer relevance analysis in Figure~\ref{fig:motivation}, we introduce a fixed layer budget $m$, where $1\leq m\leq n$, to define the bound of world model layers considered for representation aggregation.
Specifically, we use the first $m$ layers to retain informative representations while avoiding unnecessary aggregation from the remaining $n-m$ layers.
During training, all $n$ world model layers are executed for video co-training, while only the representations from the first $m$ layers are used for action prediction.
During inference, only the first $m$ layers are executed, and their key and value representations are cached for reuse across action denoising steps.

Within this bounded range, we estimate aggregation weights for different world model layers based on the current action denoising state.
Let $a_j^t$ denote the noisy action latent of the $j$th action in the action chunk at denoising step $t$, where $a_{1:H}^t=(a_1^t,\ldots,a_H^t)$.
We first average the noisy action latents across all $H$ actions to obtain a summary vector and compute an aggregation score $\alpha_{t,i}$ for each participating world model layer:
\begin{equation}
    \alpha_{t,i}=
    \frac{
    \left(
    W_q\left(\frac{1}{H}\sum_{j=1}^{H}a_j^t\right)
    \right)^{\mathsf T}e_i
    }{\sqrt{r}},
\end{equation}
where $W_q$ is a learned projection matrix, $e_i$ is a learned routing vector for layer $i$, and $r$ is the dimension of the projected query and routing vectors.
Based on these scores, we estimate the denoising-state-dependent aggregation weights as:
\begin{equation}
    w_{t,i}=
    \frac{\exp(\alpha_{t,i})}
    {\sum_{k=1}^{m}\exp(\alpha_{t,k})},
    \label{eq:action_weights}
\end{equation}
These weights allow the contributions of different world model layers to vary according to the current action denoising state (as illustrated in Figures~\ref{fig:motivation}(c) and (d)).
To further enhance representation aggregation, we introduce learnable layer-wise weights that capture the relative contributions of individual world model layers.
Specifically, we assign a learnable logit $\beta_i$ to each participating layer and reformulate Eq.~\eqref{eq:action_weights} to obtain the final fusion weights:
\begin{equation}
\begin{aligned}
    w_{t,i} =
    \frac{\exp(\gamma_\alpha\alpha_{t,i}+\gamma_\beta\beta_i)}
    {\sum_{k=1}^{m}\exp(\gamma_\alpha\alpha_{t,k}+\gamma_\beta\beta_k)},
\end{aligned}
\label{eq:aggregation_weights}
\end{equation}
where $\gamma_\alpha$ and $\gamma_\beta$ are weighting coefficients that control the influence of denoising-state-dependent aggregation and layer-wise aggregation, respectively.
In this way, the final fusion weights incorporate both the current action denoising state and the learned layer-wise aggregation preferences.
Aggregation is performed independently for each attention head; for clarity, we omit the attention-head index throughout the formulations, while $\gamma$ is shared across heads.

Using the resulting fusion weights, we aggregate the keys and values as:
\begin{equation}
    \bar{K}_t=\sum_{i=1}^{m}w_{t,i}K_i,
    \qquad
    \bar{V}_t=\sum_{i=1}^{m}w_{t,i}V_i.
    \label{eq:fused_kv}
\end{equation}
Here, $z_o^t=(\bar{K}_t,\bar{V}_t)$ denotes the aggregated world representations at denoising step $t$.
The same fusion weights are applied to keys and values and shared across spatial token positions, preserving their spatial arrangement and key-value correspondence.
Following~\cite{ma2026faster}, we further apply RMS normalization and positional encoding to the fused keys.

\vspace{-0.3cm}
\paragraph{Action Prediction.}
The action head incorporates the aggregated world representations into action prediction.
The attention output is computed as:
\begin{equation}
    O_t^a=
    \operatorname{Softmax}\left(
    \frac{
    Q_t^a[\bar{K}_t;K_t^a]^{\mathsf T}
    }{\sqrt{d_{\mathrm{head}}}}
    \right)
    [\bar{V}_t;V_t^a],
\end{equation}
where $O_t^a$ denotes the attention output; $Q_t^a$, $K_t^a$, and $V_t^a$ denote the query, key, and value representations of the action head at denoising step $t$, respectively; $d_{\mathrm{head}}$ is the attention-head dimension; and $[\cdot;\cdot]$ denotes concatenation.
The output $O_t^a$ is then processed by the remaining action head operations, including an output projection layer and a feed-forward layer, to predict the action velocity associated with the noisy action sequence $a_{1:H}^t$.
The sampler uses the predicted velocity to update $a_{1:H}^t$ to $a_{1:H}^{t-1}$.

\begin{figure}[t!]
\centering
\includegraphics[width=1.0\linewidth]{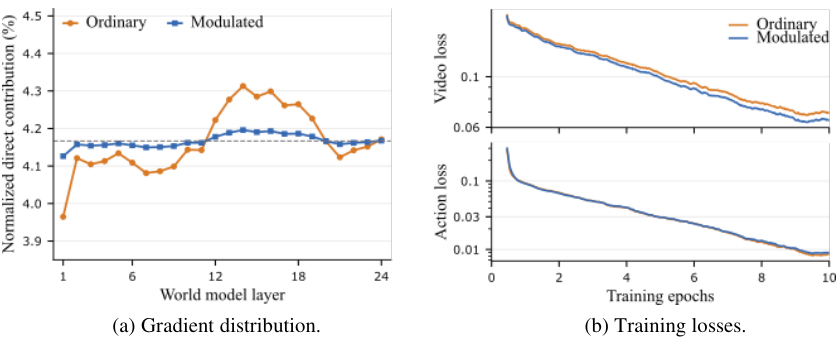}
\caption{Motivation for world model gradient modulation in TurboWAM.
(a) Action-supervision gradient distribution across world model layers before and after applying world model gradient modulation.
(b) Video and action training losses curves.
}
\label{fig:motivation_2}
\end{figure}

\subsection{World Model Gradient Modulation}
\label{sec:world_model_gradient_modulation}

Following~\cite{yuan2026fast,ma2026faster}, we jointly optimize the video and action flow-matching objectives as:
\begin{equation}
    \mathcal{L}=\mathcal{L}_v+\mathcal{L}_a,
    \label{eq:twam_loss}
\end{equation}
where $\mathcal{L}_v$ and $\mathcal{L}_a$ denote the video and action flow-matching losses, respectively.
During ordinary backpropagation, action supervision is propagated through the weighted aggregation to the participating world model layers.
Since the representations are combined using different fusion weights, the resulting direct action-supervision gradients are imbalanced across layers (see Figure~\ref{fig:motivation_2}(a)).
In particular, layers with smaller fusion weights receive weaker direct gradients, potentially limiting their optimization for world representation learning.
This imbalance may also affect the co-training of the video world model and action prediction (see Figure~\ref{fig:motivation_2}(b)), suggesting a trade-off between action fitting and world model learning.
To resolve this problem, we propose a world model gradient modulation mechanism that decomposes action-supervision gradients into layer-shared and layer-specific residual components.

Specifically, let $X_i=(K_i,V_i)$ represent the key/value features extracted from the $i$th participating world model layer, and let $g_t={\partial\mathcal{L}_a}/{\partial z_o^t}$ denote the gradient of the action loss with respect to the aggregated world representations $z_o^t$ at denoising step $t$.
Since the fusion weights depend on the noisy action latents and learned routing parameters, the direct gradient contribution from the action loss to the K/V representations of the $i$th layer through the fusion operation is:
\begin{equation}
    G_{t,i}
    =
    \left.
    \frac{\partial\mathcal{L}_a}{\partial X_i}
    \right|_{\mathrm{fusion},t}
    =
    w_{t,i}g_t.
    \label{eq:direct_gradient}
\end{equation}
Thus, the direct action-supervision gradient received by each layer is proportional to its forward fusion weight, potentially limiting the optimization of world representations from layers with smaller weights.
To regulate the gradient distribution across layers, we decompose the direct gradients into a layer-shared component and layer-specific residual components, and modulate their contributions as:
\begin{equation}
\begin{aligned}
    G_{t,i}' &=
    \mu_{\mathrm{com}}G_{t,\mathrm{com}}
    +\mu_{\mathrm{res}}G_{t,\mathrm{res},i},\\
    G_{t,\mathrm{com}} &=
    \frac{1}{m}\sum_{j=1}^{m}G_{t,j},
    \quad G_{t,\mathrm{res},i} =
    G_{t,i}-G_{t,\mathrm{com}},
\end{aligned}
\label{eq:gradient_decomposition}
\end{equation}
where $\mu_{\mathrm{com}}$ and $\mu_{\mathrm{res}}$ control the contributions of the layer-shared and layer-specific residual components, respectively.
The layer-shared component provides common action supervision across the participating layers, while the residual component preserves layer-specific gradient information.
The residual component is scaled by $\mu_{\mathrm{res}}$, allowing its contribution to be controlled relative to the shared component.
Combining Eqs.~\eqref{eq:gradient_decomposition} and~\eqref{eq:direct_gradient}, we obtain:
\begin{equation}
    G_{t,i}'=
    \mu_{\mathrm{com}}
    \left[
    \rho w_{t,i}+\frac{1-\rho}{m}
    \right]g_t,
    \qquad
    \rho=\frac{\mu_{\mathrm{res}}}{\mu_{\mathrm{com}}},
    \label{eq:backward_derivation}
\end{equation}
where $\mu_{\mathrm{com}}>0$.
For $0\leq\rho\leq1$, the resulting layer-wise gradient weights interpolate between the learned fusion distribution and a uniform distribution over the $m$ participating layers, promoting more balanced direct action supervision while retaining layer-specific information.
Importantly, gradient modulation is applied exclusively to the action-loss gradients propagated through the aggregation operation to the selected world model representations during training.
It does not change the forward aggregation or action prediction process and introduces no additional inference computation.
The video loss and gradients to the remaining model components follow ordinary backpropagation.

\begin{table*}[t]
    \centering
    \caption{Comparison with state-of-the-art methods on LIBERO and RoboTwin 2.0.
    EPT: Embodied Pre-Training.    
    $^{\dagger}$Implementation based on the arXiv:2608.02365v1 version.
    $^\ddagger$VLA results are for reference only.
    TurboWAM* denotes the compiled version of TurboWAM accelerated using \texttt{torch.compile} with CUDA Graph replay, with compilation and graph capture performed during warm-up before model inference measurement.
    }
    \label{tab:main_results}
    % \small
    \resizebox{1.0\textwidth}{!}{%
    \begin{tabular}{c|c|c|cccccc|cccc}
        \toprule
        \multirow{2}{*}{Type}
        & \multirow{2}{*}{Method}
        & \multirow{2}{*}{EPT}
        & \multicolumn{6}{c|}{LIBERO}
        & \multicolumn{4}{c}{RoboTwin 2.0} \\
        \cmidrule(lr){4-9}
        \cmidrule(lr){10-13}
        & & 
        & Spatial$\uparrow$
        & Object$\uparrow$
        & Goal$\uparrow$
        & Long$\uparrow$
        & Avg$\uparrow$
        & Latency$\downarrow$(ms)
        & Clean$\uparrow$
        & Rand$\uparrow$
        & Avg$\uparrow$
        & Latency$\downarrow$(ms) \\
        \midrule

        \multirow{2}{*}{VLA$^\ddagger$}
        & X-VLA~\cite{zheng2026x} & \cmark
        & 98.2 & 98.6 & 97.8 & 97.8 & 98.1
        & 113.3
        & 72.9 & 72.8 & 72.9 & 125.0 \\

        & $\pi_{0.5}$~\cite{black2025pi_05} & \cmark
        & 98.8 & 98.2 & 98.0 & 92.4 & 96.9
        & 42.0
        & 82.7 & 76.8 & 79.8 & 36.7 \\

        \midrule

        \multirow{8}{*}{WAM}
        & CosmosPolicy~\cite{kim2026cosmos} & \cmark
        & 98.1 & 100.0 & 98.2 & 97.6
        & \cellcolor{tabsecond}{98.5}
        & 296.9
        & -- & -- & -- & -- \\

        & Motus~\cite{bi2026motus} & \cmark
        & 96.8 & 99.8 & 96.6 & 97.6 & 97.7
        & 1555.3
        & 88.7 & 87.0 & 87.8 & 1534.7 \\

        & LingBot-VA~\cite{li2026causal} & \cmark
        & 98.5 & 99.6 & 97.2 & 98.5
        & \cellcolor{tabsecond}{98.5}
        & 3032.3
        & 92.9 & 91.5
        & \cellcolor{tabfirst}{92.2}
        & 3328.0 \\

        & Efficient-WAM~\cite{li2026efficient} & \xmark
        & -- & -- & -- & -- & --
        & --
        & 86.7 & 85.7 & 86.2 & 329.2 \\

        & Light-WAM~\cite{li2026light} & \xmark
        & 98.2 & 99.6 & 97.8 & 93.0 & 97.2
        & \cellcolor{tabthird}{118.5}
        & 76.4 & 76.3 & 76.4
        & \cellcolor{tabthird}{116.9} \\

        & Faster-WAM$^{\dagger}$~\cite{ma2026faster} & \xmark
        & 98.0 & 99.8 & 98.6 & 96.4
        & \cellcolor{tabthird}{98.2}
        & \cellcolor{tabsecond}{76.1}
        & 89.6 & 89.5 & 89.6
        & \cellcolor{tabsecond}{77.1} \\

        & Fast-WAM~\cite{yuan2026fast} & \xmark
        & 98.2 & 100.0 & 97.0 & 95.2 & 97.6
        & 336.7
        & 91.9 & 91.8
        & \cellcolor{tabsecond}{91.8}
        & 437.2 \\

        & TurboWAM (ours) & \xmark
        & 99.4 & 99.8 & 98.6 & 98.0
        & \cellcolor{tabfirst}{99.0}
        & \cellcolor{tabfirst}{49.7}
        & 90.4 & 90.5
        & \cellcolor{tabthird}{90.5}
        & \cellcolor{tabfirst}{47.1} \\

        \rowcolor{gray!20}
        & TurboWAM* (ours) & \xmark
        & 99.0 & 100.0 & 99.6 & 97.2
        & 99.0
        & 16.8
        & 90.6 & 90.0
        & 90.3
        & 21.8 \\

        \bottomrule
    \end{tabular}%
    }
\end{table*}
\begin{table*}[t]
    \centering
    \caption{Comparison with state-of-the-art methods on LIBERO-Plus across seven perturbation categories.
    TurboWAM$^+$ denotes results obtained with corrected language prompts following https://github.com/sylvestf/LIBERO-plus/issues/64.
    }
    \label{tab:libero_plus}
    % \small
    \resizebox{1.0\textwidth}{!}{%
    \begin{tabular}{c|c|c|ccccccccc}
        \toprule
        \multirow{2}{*}{Type}
        & \multirow{2}{*}{Method}
        & \multirow{2}{*}{EPT}
        & \multicolumn{9}{c}{LIBERO-Plus} \\
        \cmidrule(lr){4-12}
        & &
        & Camera$\uparrow$
        & Robot$\uparrow$
        & Language$\uparrow$
        & Light$\uparrow$
        & Background$\uparrow$
        & Sensor$\uparrow$
        & Object$\uparrow$
        & Avg$\uparrow$
        & Latency (ms)$\downarrow$ \\
        \midrule

        \multirow{2}{*}{VLA$^\ddagger$}
        & X-VLA~\cite{zheng2026x} & \cmark
        & 23.4 & 89.7 & 75.7 & 88.2
        & 96.0 & 62.7 & 71.8
        & 71.4
        & 128.9 \\

        & $\pi_{0.5}$~\cite{black2025pi_05} & \cmark
        & 75.4 & 77.5 & 85.6 & 96.9
        & 94.6 & 89.7 & 85.7
        & 85.7
        & 42.4 \\

        \midrule

        \multirow{6}{*}{WAM}
        & CosmosPolicy~\cite{kim2026cosmos} & \cmark
        & 75.8 & 63.3 & 81.7 & 96.5
        & 88.9 & 92.7 & 82.2
        & \cellcolor{tabsecond}{82.2}
        & 323.8\\

        & LingBot-VA~\cite{li2026causal} & \cmark
        & 40.9 & 83.0 & 86.4 & 82.3
        & 53.1 & 64.4 & 76.2
        & 69.5
        & 3757.0 \\

        & Faster-WAM$^{*}$~\cite{ma2026faster} & \xmark
        & 71.6 & 54.0 & 89.6 & 93.6
        & 55.3 & 81.6 & 80.2
        & 75.3
        & \cellcolor{tabsecond}{84.9} \\

        & Fast-WAM~\cite{yuan2026fast} & \xmark
        & 16.4 & 44.5 & 68.9 & 78.2
        & 53.7 & 37.7 & 60.7
        & 51.5
        & \cellcolor{tabthird}{351.9} \\

        & TurboWAM (ours) & \xmark
        & 69.5 & 65.2 & 93.0 & 93.7
        & 63.9 & 78.9 & 82.6
        & \cellcolor{tabthird}{78.1}
        & \cellcolor{tabfirst}{50.1} \\

        & TurboWAM$^+$(ours) & \xmark
        & 81.0 & 74.5 & 93.0 & 93.2
        & 65.4 & 89.1 & 83.7
        & \cellcolor{tabfirst}{83.3}
        & \cellcolor{tabfirst}{50.1} \\

        \rowcolor{gray!20}
        & TurboWAM$^+$*(ours) & \xmark
        & 81.1 & 73.6 & 92.9 & 93.8
        & 65.2 & 87.7 & 83.5
        & 82.9
        & 17.0 \\

        \bottomrule
    \end{tabular}%
    }
\end{table*}

\section{Experiments}
\label{sec:experiments}

\subsection{Experimental Setup}
We evaluate TurboWAM on three widely used robot manipulation benchmarks, namely LIBERO~\cite{liu2023libero}, LIBERO-Plus~\cite{fei2026libero}, and RoboTwin2.0~\cite{chen2025robotwin}, and additionally conduct experiments on a real-world robot platform involving several manipulation tasks.
LIBERO consists of four task suites, namely Spatial, Object, Goal, and Long, each containing 10 tasks with 500 demonstrations.
We train TurboWAM on all four suites and report task success rates over 2,000 evaluation trials across 40 tasks.
LIBERO-Plus extends LIBERO by introducing seven categories of perturbations, including camera viewpoints, robot initial states, language instructions, lighting conditions, background textures, sensor noise, and object layouts, to evaluate the robustness and generalization capabilities of robot policies.
We evaluate the zero-shot generalization of TurboWAM on 10,030 LIBERO-Plus instances by directly applying the models trained on the original LIBERO suites without further adaptation.
We report task success rates across all seven perturbation categories.
RoboTwin2.0 is a bimanual robot manipulation benchmark comprising 50 tasks.
Following~\cite{yuan2026fast,ma2026faster}, we train TurboWAM using 2,500 demonstrations from clean scenes and 25,000 demonstrations from randomized scenes across all 50 tasks.
We report task success rates over 100 evaluation trials per task under both clean and randomized conditions.
For real-world robot experiments, details of the experimental setup are provided in the supplementary material.

\subsection{Implementation Details}
We implement TurboWAM using Python and PyTorch.
We adopt the VAE and video world model from the pretrained Wan2.2-5B~\cite{wan2025wan} and employ T5~\cite{raffel2020exploring} as the text encoder.
A linear projection layer is used to encode proprioceptive states into the same feature dimension as the language embeddings before they are fed into the video world model.
We set the action chunk horizon to $H{=}32$, the layer budget to $m{=}24$, and the routing dimension to $r{=}16$.
For gradient modulation, we set $\mu_{\mathrm{com}}{=}0.25$ and $\mu_{\mathrm{res}}{=}0.05$.
We optimize the model using AdamW~\cite{loshchilov2017decoupled} with a cosine annealing learning rate schedule and linear warm-up, where the maximum learning rate is set to $10^{-4}$.
We train TurboWAM for 10 epochs on LIBERO and 5 epochs on RoboTwin2.0 with 4 GPUs.
Inference latency is measured with an action chunk horizon of 32, using cached instruction embeddings.
The inference latency of all compared methods is measured on the same single data-center GPU with approximately 140 GB of memory under identical environmental settings.

\subsection{Comparison with State-of-the-Art Methods}

\paragraph{Compared Methods.}
We compare TurboWAM against state-of-the-art WAM methods, including Motus\cite{bi2026motus}, LingBot-VA\cite{li2026causal}, Cosmos Policy\cite{kim2026cosmos}, Efficient-WAM\cite{li2026efficient}, Light-WAM\cite{li2026light}, Faster-WAM\cite{ma2026faster}, and Fast-WAM\cite{yuan2026fast}.
We also include representative VLA-based methods (X-VLA\cite{zheng2026x} and $\pi_{0.5}$\cite{black2025pi_05}) for reference.

\begin{table*}[t]
    \centering
    \caption{Ablation study of the proposed TurboWAM components.}
    \label{tab:ablation_component}
    \small
    \begin{tabular}{c|cc|cc|cc}
        \toprule
        \multirow{2}{*}{Variant}
        & \multicolumn{2}{c|}{World representation aggregation}
        & \multicolumn{2}{c|}{World model gradient modulation}
        & \multicolumn{2}{c}{Avg. success rate $\uparrow$} \\
        \cmidrule(lr){2-3}
        \cmidrule(lr){4-5}
        \cmidrule(lr){6-7}
        & Denoising-state-dependent
        & Layer-wise
        & Layer-shared
        & Layer-specific
        & LIBERO
        & LIBERO-Plus \\
        \midrule
        Full & \cmark & \cmark & \cmark & \cmark & \textbf{99.0} & \textbf{78.1} \\        
        V1 & \cmark & \xmark & \cmark & \cmark & \textbf{99.0} & 77.6 \\
        V2 & \xmark & \xmark & \cmark & \cmark & 98.5 & 75.9 \\
        V3 & \cmark & \cmark & \cmark & \xmark & 98.5 & 75.7 \\
        V4 & \cmark & \cmark & \xmark & \xmark & 97.9 & 69.2 \\        
        \bottomrule
    \end{tabular}
\end{table*}

\begin{table*}[t]
\centering
\caption{Performance on real-world robot manipulation experiments. Speedup is relative to Fast-WAM.}
\label{tab:real_robot}
\resizebox{0.95\textwidth}{!}{
\begin{tabular}{@{}l ccc ccc ccc ccc cc@{}}
\hline
\multirow{2}{*}{Method}
& \multicolumn{3}{c}{Pick-and-Place}
& \multicolumn{3}{c}{Sequential}
& \multicolumn{3}{c}{Sorting}
& \multicolumn{3}{c}{Average}
& \multirow{2}{*}{Latency(ms)$\downarrow$}
& \multirow{2}{*}{Speedup$\uparrow$} \\
\cline{2-13}
& SR$\uparrow$ & PS$\uparrow$ & PQ$\downarrow$
& SR$\uparrow$ & PS$\uparrow$ & PQ$\downarrow$
& SR$\uparrow$ & PS$\uparrow$ & PQ$\downarrow$
& SR$\uparrow$ & PS$\uparrow$ & PQ$\downarrow$
& & \\
\hline
Fast-WAM~\cite{yuan2026fast}
& 85.0 & 85.0 & 38.2
& {60.0} & 67.5 & 56.4
& 65.0 & 70.0 & 73.8
& 70.0 & 74.2 & 56.1
& 358.8 & 1.0$\times$ \\

Faster-WAM~\cite{ma2026faster}
& 85.0 & 85.0 & 38.5
& 50.0 & 60.0 & 54.9
& 65.0 & 72.5 & 71.1
& 66.7 & 72.5 & 54.8
& 87.6 & 4.1$\times$ \\

TurboWAM (ours)
& {90.0} & {90.0} & {38.1}
& {60.0} & {70.0} & {54.7}
& {70.0} & {75.0} & {70.2}
& \textbf{73.3} & \textbf{78.3} & \textbf{54.3}
& \textbf{66.5} & \textbf{5.4}$\times$ \\
\hline
\end{tabular}
}
\end{table*}

\vspace{-0.3cm}
\paragraph{Result Analysis.}
As shown in Table~\ref{tab:main_results}, TurboWAM achieves competitive task success rates with substantially reduced inference latency on both LIBERO and RoboTwin2.0.
Specifically, on LIBERO, TurboWAM achieves the best average success rate of 99.0\% and provides 6.8$\times$/1.5$\times$ inference speedups over Fast-WAM/Faster-WAM.
On RoboTwin2.0, TurboWAM maintains strong task performance, achieving an average success rate of 90.5\% with inference speedups of 9.3$\times$/1.6$\times$ over Fast-WAM/Faster-WAM.
We further evaluate the zero-shot generalization of TurboWAM on LIBERO-Plus.
As shown in Table~\ref{tab:libero_plus}, TurboWAM achieves competitive performance across the seven perturbation categories compared with state-of-the-art methods.
Notably, these results are obtained by directly evaluating models trained on the original LIBERO suites without any additional adaptation.
This demonstrates the robustness and generalization capability of TurboWAM under diverse perturbations.
In addition, as shown in Table~\ref{tab:main_results}, the compiled version TurboWAM* further reduces inference latency to 16.8 ms on LIBERO and 21.8 ms on RoboTwin2.0 while maintaining comparable task success rates.

\subsection{Ablation Study}

\paragraph{World Representation Aggregation.}
In Table~\ref{tab:ablation_component}, we compare the default world representation aggregation mechanism (Full) with two variants, including V1 that removes learnable layer-wise weights and V2 that uses direct averaging of the world representations.
We can observe that the full aggregation consistently outperforms these alternative strategies.
Directly averaging world representations results in substantially worse performance, while using only denoising-state-dependent weights leads to slight performance drops on LIBERO-Plus.

\vspace{-0.3cm}
\paragraph{World Model Gradient Modulation.}
In Table~\ref{tab:ablation_component}, we compare the default world model gradient modulation mechanism (Full) with two variants, including V3 that uses only the layer-shared component and V4 that uses ordinary action supervision without gradient modulation.
It can be seen that the proposed gradient modulation consistently achieves better performance than the alternative strategies.
Since we use a bounded set of layers for world representation aggregation, directly applying ordinary action supervision without gradient modulation may lead to suboptimal updates for the selected layers, thereby degrading overall performance.
Using only the layer-shared component may fail to capture layer-specific nuances in the world representations, resulting in suboptimal optimization.
We additionally evaluate a gradient-insulation variant inspired by~\cite{driess2026knowledge}, which blocks continuous action-loss gradients into the world model and instead provides discrete-action supervision through a lightweight decoder.
This variant achieves substantially lower success rates of 93.9\% on LIBERO and 56.6\% on LIBERO-Plus.

\section{Real-World Robot Experiments}
\label{sec:suppl_real_world}

We additionally evaluate whether the favorable trade-off between inference latency and task success of TurboWAM transfers to physical robot control.

\vspace{-0.3cm}
\paragraph{Evaluation Protocol.}
The real-world robot manipulation tasks are performed on a robot platform with a single robot arm and two cameras.
We follow the LIBERO experimental setup~\cite{liu2023libero} for image processing, policy learning and evaluation.
We design three tasks with different manipulation requirements, including \emph{Precise Pick-and-Place}, \emph{Sequential Stacking and Placement}, and \emph{Language-Guided Sorting}.
For each task, we collect 100 successful demonstrations for training and use 20 test scenes for evaluation.
The scenes vary the placement of the task objects across the workspace and include up to several distractors, such as fruits, toy food, bowls, trays, cups of other colors, a pen holder, markers, cans and cartons.
We report the task \emph{success rate} (SR), the \emph{progress score} (PS) and the \emph{policy queries} (PQ).
Please refer to the \emph{supplementary material} for the details of the experimental setup and evaluation metrics.

\vspace{-0.3cm}
\paragraph{Result Analysis.}
We report the results in Table~\ref{tab:real_robot}.
Overall, TurboWAM achieves the best performance across all tasks.
Specifically, on the Precise Pick-and-Place task, all methods achieve high success rates, and TurboWAM attains the highest success rate and the lowest query count, indicating both effectiveness and efficiency.
On the Sequential Stacking and Placement task, TurboWAM matches or surpasses the other methods in success rate and progress score while maintaining a lower query count.
For the Language-Guided Sorting task, TurboWAM achieves the best success rate and progress score.
Under the same real-world conditions, TurboWAM achieves inference speedup of 5.4$\times$ over Fast-WAM.

\section{Conclusion}
\label{sec:conclusion}

In this work, we present TurboWAM, a novel fast world action model that reduces inference latency while maintaining competitive task success rates.
TurboWAM introduces a world representation aggregation mechanism to maintain informative world representations for action prediction while reducing computational overhead.
Furthermore, TurboWAM employs a world model gradient modulation mechanism to optimize world representations more effectively during action learning.
Extensive experiments show that TurboWAM achieves significantly lower inference latency while maintaining competitive task success rates compared with state-of-the-art methods.

\paragraph{Limitations and Future Work.}
Despite its promising performance, TurboWAM still has several limitations.
First, it may face challenges in scenarios requiring complex spatial reasoning and intricate object interactions.
Incorporating geometric priors from spatial foundation models~\cite{wang2025vggt,huang2026turbovggt} or semantic knowledge from vision-language models~\cite{bai2025qwen3} could be promising directions to address these challenges.
Second, further improving task success rates while preserving low inference latency remains an important research direction.
Future work will explore these directions to further enhance the robustness and efficiency of TurboWAM.

{
    \small
    \bibliographystyle{ieeenat_fullname}
    \bibliography{ref}
}

\end{document}